\documentclass{article}
\usepackage{graphicx} 
\usepackage{amsmath}
\usepackage{booktabs}
\usepackage{multicol}
\usepackage[a4paper, total={6in, 8in}]{geometry}
\usepackage{hyperref}
\usepackage{subcaption}

\title{Efficient Video Sampling: Pruning Temporally Redundant Tokens for Faster VLM Inference}

\author{%
  Natan Bagrov, Eugene Khvedchenia, Borys Tymchenko\\[1.5ex]
  Shay Aharon, Lior Kadoch, Tomer Keren, Ofri Masad\\[1.5ex]
  Yonatan Geifman, Ran Zilberstein\\[1.5ex]
  Tuomas Rintamaki, Matthieu Le, Andrew Tao\\[3ex]
  \texttt{Correspondence: evs-paper@nvidia.com}
}

\date{} 

\begin{document}

\maketitle

\begin{abstract}
Vision-language models (VLMs) have recently expanded from static image understanding to video reasoning, but their scalability is fundamentally limited by the quadratic cost of processing dense frame sequences. Long videos often exceed the token budget of modern language models, leading to severe context limitations and latency issues. We introduce \textbf{Efficient Video Sampling (EVS)}, a simple, plug-and-play method for reducing token redundancy in videos by identifying and pruning \emph{temporally static patches} -- spatial regions that remain unchanged across consecutive frames. EVS preserves positional identity, requires no architectural changes or retraining. We show that EVS substantially reduces token count while maintaining semantic fidelity, enabling faster inference and longer input sequences. Applied at inference time, EVS reduces large language model (LLM) time-to-first-token (TTFT) by up to 4$\times$ with minimal accuracy loss. When combined with an uptraining phase using stochastic pruning rates, EVS yields models that are robust to varying compression levels and retain full performance under aggressive pruning. Extensive experiments demonstrate that EVS consistently improves efficiency--accuracy trade-offs, unlocking scalable video-language understanding without sacrificing quality.
\end{abstract}

\section{Introduction}
\label{sec:intro}

Vision-language models (VLMs) have advanced rapidly \cite{alayrac2022flamingo, liu2023visual, lin2024vila, chen2024internvl}, demonstrating strong capabilities in understanding and reasoning over visual and textual modalities. While early models focused on static images, recent work has extended them to videos \cite{Maaz2023VideoChatGPT, zhang2023video}, enabling temporal understanding across frame sequences. However, this introduces a major bottleneck: videos produce large amounts of visual tokens. For example, a two-minute video at 24 FPS produces more than two million vision tokens, far beyond the effective context length of most language models, which typically ranges from 4K to 128K tokens \cite{hong2025context}. This becomes especially problematic for longer videos, which are common in real-world applications such as surveillance, instructional content, and robot learning.

In addition to context constraints, inference latency increases with the number of tokens. Even with extended context windows, handling long token sequences imposes heavy computational loads. Reducing token count is therefore essential, not only to fit within context limits but also to enable efficient, timely video processing.

A common property of real-world video is high temporal redundancy: consecutive frames often show minimal change \cite{solari1997digital, wiegand2003overview}. It is typical in CCTV footage, traffic cameras, and long recordings where static backgrounds dominate. For example, a hallway camera may capture hundreds of nearly identical frames. Processing each frame fully incurs high computational and memory costs, resulting in little new information.

To address this, we propose \textbf{Efficient Video Sampling (EVS)}, a simple, plug-and-play method that leverages temporal redundancy. EVS detects highly similar visual patches across consecutive frames and compresses them into a single token. We target ``static patches''—regions at the same spatial location across frames that remain unchanged. These are common and can be safely aggregated without sacrificing semantic or temporal fidelity. EVS reduces token count, accelerates training and inference, and enables longer video processing—all without adding parameters or modifying model architecture.

Our contributions are as follows: (1) We present EVS, a plug-and-play token compression technique that leverages temporal redundancy to reduce input size without altering model architecture or introducing learnable components; (2) We demonstrate that EVS enables scalable processing of long video sequences within limited compute budgets, making full-video understanding feasible in practice; and (3) we show that EVS reduces LLM time-to-first-token (TTFT) and KV-cache memory by up to 4$\times$ (Pruning rate $q=0.75$) without sacrificing accuracy, improving both training and inference efficiency across a range of real-world video scenarios.

\begin{figure}[t]
    \centering
    \includegraphics[width=\linewidth]{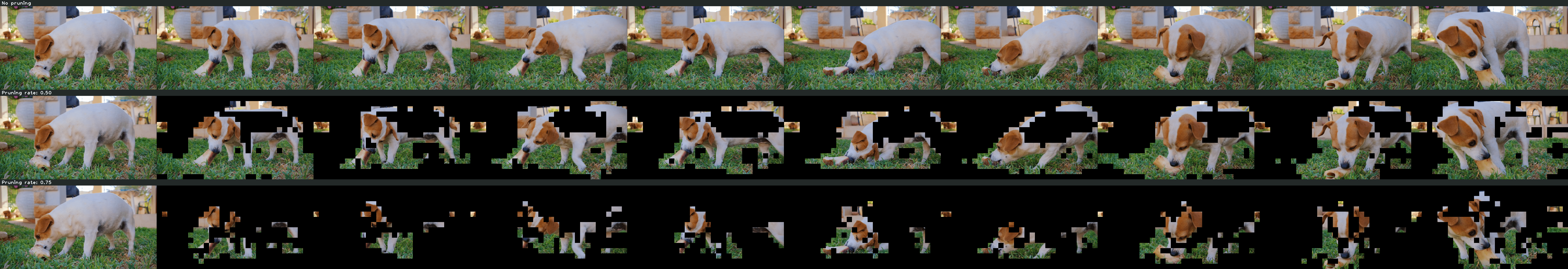}
    \caption{
        \textbf{Efficient Video Sampling: Pruning Static Patches.}
        This example video illustrates the key idea behind our Efficient Video Sampling (EVS) method.
        The top row displays a sequence of original video frames that depict a dog playing with a bone.
        The second and third rows shows the same sequence after EVS with a sequence-level pruning ratio of $q=50\%$ and $q=75\%$ respectively. \textbf{Importantly, EVS does not prune each frame uniformly: more dynamic frames are pruned less aggressively, while static frames are pruned more heavily}.
        EVS selectively retains only the most dynamic and informative patches—such as the dog's head playing with the bone.
        This enables significant token reduction without sacrificing semantic content, making EVS particularly effective for videos with sparse motion.
    }
    \label{fig:evs_static_pruning}
\end{figure}

\section{Related Work}
\label{sec:related}

Existing approaches for reducing token redundancy often rely on learned mechanisms or additional processing steps. For instance, \textbf{LongVU} \cite{shen2024longvuspatiotemporaladaptivecompression} introduces an extra module to select keyframes from long videos, adding extra parameters and complexity; however, it treats each frame as a whole, where it can be either used or skipped.

\textbf{VilaMP} \cite{cheng2025scalingvideolanguagemodels10k} addresses this limitation through hierarchical differential distillation, where learned modules are used to select keyframes and extract query-relevant information from surrounding non-keyframes. Rather than modeling patch-to-next-patch transitions uniformly across the sequence, it emphasizes patch-to-keyframe interactions, prioritizing semantically meaningful frames as anchors for temporal reasoning.  

Token Merging (\textbf{ToMe}) \cite{bolya2023tokenmergingvitfaster} reduces token counts at inference by merging similar tokens based on embedding similarity, which avoids new parameters. However, it is operating at the spatial level, ignoring temporal priors.

\textbf{NVILA} \cite{liu2025nvilaefficientfrontiervisual} introduces the ``scale-then-compress’’ paradigm, where temporal resolution is first increased to enhance accuracy, followed by token compression to regain efficiency. It partitions frames into groups and performs token averaging within each group, ignoring temporal localization and dynamic adaptation to scene changes.

\textbf{SparseVLM} \cite{zhang2025sparsevlmvisualtokensparsification} proposes a training-free, text-guided sparsification that scores visual tokens via attention maps and recycles pruned information into compact representations. While SparseVLM achieves higher compression on short clips, it requires access to attention matrices and introduces an additional computational cost associated with token recycling.

\textbf{LLaVA-Mini} \cite{zhang2025llavaminiefficientimagevideo} introduces a \emph{modality pre-fusion} module that merges visual information into the instruction text \emph{before} the language model is invoked, allowing all subsequent layers to operate on a \textbf{single} learned ``vision’’ token rather than multiple patch embeddings, not offering a smooth latency–accuracy trade-off.

Keyframe-oriented Vision Token Pruning (\textbf{KVTP}) \cite{liu2025keyframeorientedvisiontokenpruning} adaptively assigns patch-level pruning rates conditioned on a query and keeps only the most salient regions within each keyframe.

Run-Length Tokenization (\textbf{RLT}) \cite{choudhury2024dontlooktwicefaster} compresses temporally redundant tokens by encoding consecutive duplicates as a single token with a run-length indicator. Designed primarily for classification, it operates on raw pixel-level similarities, making it potentially sensitive to noise and minor variations that lack semantic relevance.

\section{Efficient Video Sampling}
\label{sec:evs}

Transformer-based large language models (LLMs) and vision-language models (VLMs) incur a quadratic cost in sequence length, making naive dense sampling of long videos prohibitive.  Efficient Video Sampling (EVS) mitigates this bottleneck by \emph{pruning nearly‐static spatio–temporal regions} while \emph{preserving the positional identity} of the retained tokens.  
Because the number of kept tokens grows sub-linearly with clip duration, EVS unlocks significantly longer temporal context without exceeding memory or latency budgets.

\begin{figure}[h]
    \centering
    \includegraphics[width=1.00\linewidth]{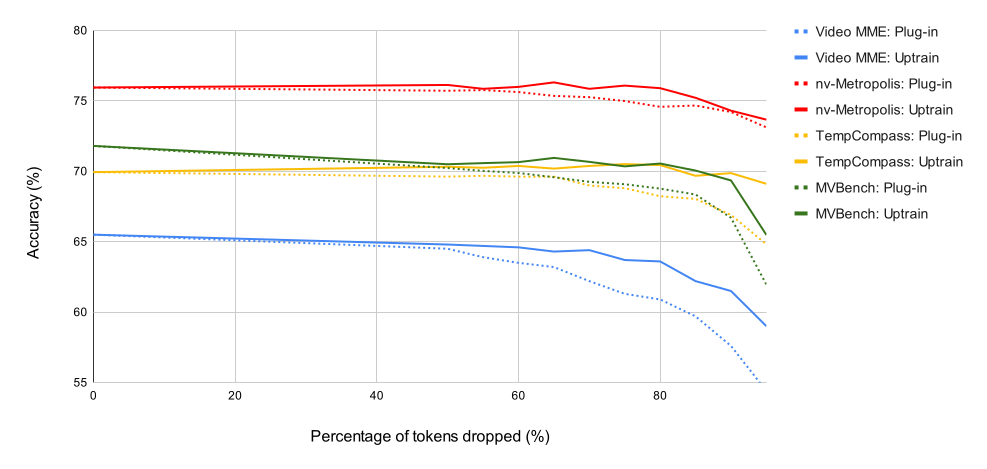}
    \caption{
        \textbf{Accuracy vs. percentage of tokens dropped by EVS.}
        This figure illustrates the trade-off between inference acceleration (measured as the percentage of tokens dropped, which directly affects TTFT) and the accuracy score, between two approaches: uptraining the model (solid lines) versus using EVS as a plug-and-play post-processing step without uptraining (dashed lines). Each color indicates a dataset. Uptraining consistently mitigates performance drops, especially at higher pruning rates.
        Notably, one can choose a pruning factor $q$ to achieve the desired tradeoff between TTFT reduction and accuracy.
        See Appendix \ref{app:additional_results} for additional details on the impact of pruning rates on the accuracy score.
    }
    \label{fig:pruning_ablation_uptrain_vs_plugin}
\end{figure}

\vspace{2pt}\noindent
\textbf{Plug-and-play design.}  
EVS is applied \emph{after} a model has been pre-trained. Since the original position encodings are forwarded unchanged for the surviving tokens, the LLM backbone treats an EVS-processed input exactly like a densely sampled one; no architectural or checkpoint modifications are required.  In practice, we observe negligible accuracy loss when EVS is activated at inference time (See Table~\ref{tab:evs-uptrain-ablation}). Intuitively, higher pruning rates result in a greater drop in accuracy. However, a short uptraining phase can help recover most of the model's accuracy while still benefiting from high pruning rates (See Figure~\ref{fig:pruning_ablation_uptrain_vs_plugin}).

\subsection{Token Selection in RGB Space}
\label{sec:evs-in-rgb-space}

In this section, we present the algorithm for selecting tokens in the RGB space.
Our RGB space token selection algorithm operates directly on raw pixel values to identify temporally redundant patches across consecutive frames. 
The algorithm computes frame-wise differences at the patch level and applies percentile-based thresholding to retain only the most dynamic regions while preserving spatial-temporal relationships.

Given a video clip \(\mathbf{X} \in \mathbf{R}^{3 \times T \times H \times W}\) and its video embedding \(\mathbf{E} \in \mathbf{R}^{C \times T \times H' \times W'}\), our goal is to compute a retention mask \(\mathbf{M} \in \{0,1\} ^ {T \times H' \times W'}\) where $H'$ and $W'$ correspond to the spatial size of embeddings \emph{after} encoding: ${W' = \left\lceil \frac{W}{\text{patch\_size}} \right\rceil}$, ${H' = \left\lceil \frac{H}{\text{patch\_size}} \right\rceil
}$.
A \textbf{patch\_size} here stands for \emph{effective} size (in pixels) of a single vision token. The effective patch size is computed as the product of the patch size from the stem block of the vision encoder and the down-sampling factor of the multimodal projector that follows the vision encoder, if applicable. 

\paragraph{EVS-mask calculation}
\begin{enumerate}
  \item Divide each frame into non-overlapping square patches of size $\mathbf{patch\_size} \times \mathbf{patch\_size}$.
  \item For every patch $p$ at time $0 < t \le T$, compute $D_{p,t}=\lVert p_t - p_{t-1}\rVert_1$, and denote $\{D_{t}\}$ as the differences of all patches between frames $t-1$ and $t$.
  \item For each frame collect $\{D_{t}\}_{t=1}^{T}$ and compute sequence-level cut-off threshold $d$ as $q$-th percentile, where $q$ is a user-selected pruning rate (e.g.\ $q=0.75$ keeps 25\,\% patches that change the most between frames).
  \item All patches in the first frame are kept unconditionally to guarantee an initial temporal anchor: ${\mathbf{M_{0}} = \mathbf{1}^{W'\times H'}}$
    \item For all patches in the consecutive frames, keep those that satisfy \(\{D_{p,t}\} \ge d\); this defines the binary mask \(\mathbf{M}_t\).
\end{enumerate}

\noindent
RGB-space selection offers significant advantages for real-time applications. Unlike embedding-space approaches, it does not require passing images through the vision encoder to identify regions for pruning, enabling low-latency pruning decisions. 
This makes it particularly suitable for streaming scenarios such as robotics, where frames arrive sequentially and immediate processing decisions are required. 
The computational savings from avoiding unnecessary encoder passes can be substantial in latency-critical applications.

\subsection{Token Selection in Embedding Space}
\label{sec:evs-in-embeddings-space}

Alternatively, Efficient Video Sampling (EVS) can be applied to \emph{post-encoder} embeddings. Let $\mathbf{E} \in \mathbf{R}^{C \times T \times H' \times W'}$ denote the embeddings of video $\mathbf{X}$, as produced by the vision encoder. We compute frame-wise cosine dissimilarity along the features' dimension $\mathbf{C}$, and apply a percentile-based thresholding mechanism to identify informative frames, analogous to the pruning strategy employed in RGB space. Likewise to the RGB-space pruning, all embeddings corresponding to the first frame of $\mathbf{X}$ are retained to provide a consistent reference.

The key distinction from RGB-space pruning lies in the choice of similarity metric. Specifically, EVS in embedding space relies on cosine similarity applied to the vision encoder's output, which offers a more semantically rich and stable representation. This is expected to yield improved robustness to minor brightness variations, sensor noise, and slight camera motion.

A comparative evaluation of embedding-space pruning versus RGB-space pruning is presented in Table~\ref{tab:evs-rgb-vs-embedding-pruning}. Notably, the embedding-based approach offers potential for integration with learnable distance metrics and adaptive, input-dependent pruning strategies. Exploration of such extensions is left for future work.

\begin{table}[h]
\centering
\begin{tabular}{@{}lcc}
\toprule
\textbf{Benchmark} & \textbf{RGB-Space} & \textbf{Embedding-Space} \\
\midrule
\textbf{Video MME} $\uparrow$ & & \\
\quad 8 frames      & 54.30 & \textbf{56.00} \\
\quad 32 frames     & 61.30 & 61.30 \\
\quad 128 frames    & 63.80 & \textbf{64.20} \\

\textbf{nv-Metropolis} $\uparrow$ & & \\
\quad 8 frames   & 73.07 & \textbf{73.76} \\
\quad 32 frames  & 74.90 & \textbf{74.99} \\

\textbf{TempCompass MCQ} $\uparrow$ & & \\
\quad 8 frames   & 67.34 & \textbf{67.59} \\
\quad 32 frames  & 68.80 & 68.80 \\

\textbf{MVBench} $\uparrow$ & & \\
\quad 8 frames   & 65.33 & \textbf{66.07} \\
\quad 32 frames  & \textbf{69.92} & 69.08 \\

\bottomrule
\end{tabular}
\caption{Comparison of RGB-space vs Embedding-space pruning. Evaluation performed with fixed rate $q=0.75$ and preserving position IDs, using the Qwen 7B model without uptraining. }
\label{tab:evs-rgb-vs-embedding-pruning}
\end{table}

\subsection{Position-Preserving Pruning}
\label{sec:position-ids}

\begin{figure}
    \centering
    
    \includegraphics[width=\linewidth]{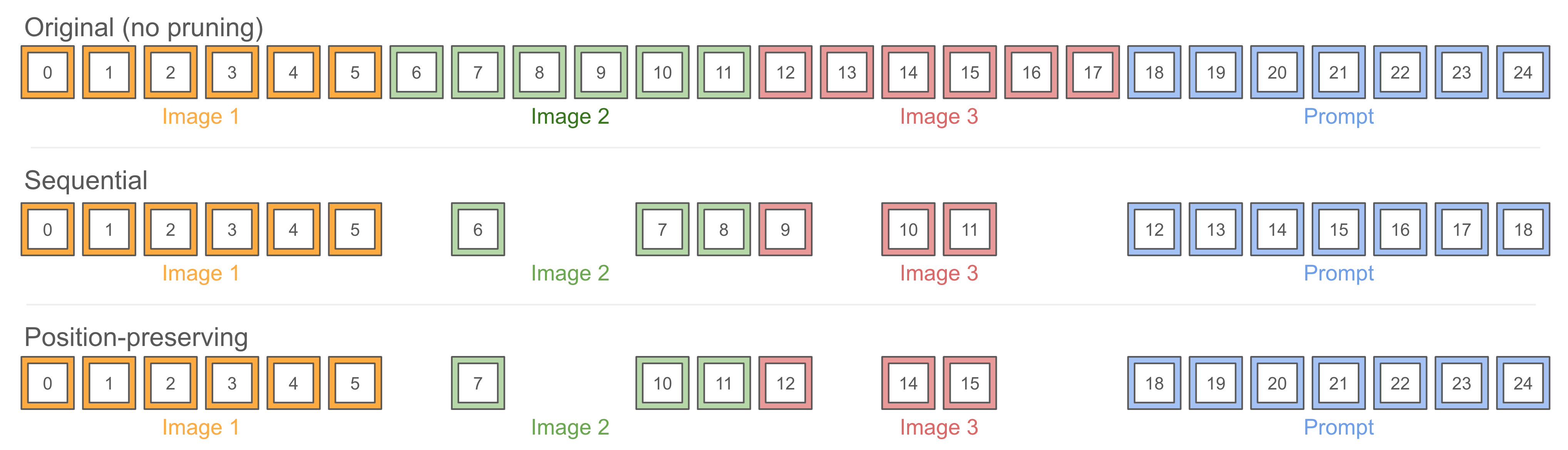}
    \caption{Visual explanation of Sequential and Position-preserving handling of pruned tokens. The top row indicates position IDs assigned to input images and text prompts without pruning. Empty cells correspond to pruned tokens. In \textbf{Sequential} method, position ids for input tokens are computed for output tokens \emph{after} pruning. In \textbf{Position-preserving} method, position IDs are computed for token positions \emph{before} pruning.}
    \label{fig:position-preserving-figure}
\end{figure}

Modern LLMs employ one or another form of positional encoding to embed location information of a token within a sentence. Naturally, pruning reduces the number of tokens we pass to LLM. We hypothesize that the original position IDs corresponding to an original (non-pruned) sequence carry important information about the spatial location of each vision patch and thus play a crucial role in preserving the model's accuracy. We call it \textbf{Position-preserving} encoding. To ablate this assumption, we implement another method, \textbf{Sequential} encoding, in which we assign monotonically increasing position IDs to the final, pruned sequence of tokens (see Figure \ref{fig:position-preserving-figure}).

For \textbf{Position-preserving} encoding, we gather both the selected embeddings and their original positional indices based on the computed retention mask $\mathbf{M}$:

\[
\mathbf{E}' = \mathrm{gather}(\mathbf{E}, \mathbf{M}),
\quad
\mathbf{P}' = \mathrm{gather}(\texttt{position\_ids}, \mathbf{M}).
\]

\noindent
The language module, therefore, receives a shorter sequence $(\mathbf{E}',\mathbf{P}')$ that is \emph{semantically and positionally consistent} with the full input.

Our ablations in Section \ref{sec:exp-position-ids} reveal an interesting insight: different strategies are preferable for plug-and-play and uptrained models. For plug-and-play, it is better to use the position IDs as-is, while for uptraining, it is advantageous to preserve them.

\subsection{Model Uptraining}

We perform a brief fine-tuning phase in which the pruning rate $q$ is sampled from a beta distribution for every mini-batch.  
The model thus learns to be invariant to a continuum of compression ratios and retains its performance whether EVS is enabled or disabled at inference.  

For newly built VLMs, Uptraining can replace standard supervised fine-tuning, producing a model that can operate at both regular, dense (EVS turned off) vision input and sparse (EVS turned on) vision input (see Table~\ref{tab:evs-uptrain-ablation}).

\subsection{Runtime flexibility}

As shown in Figure \ref{fig:pruning_ablation_uptrain_vs_plugin}, the pruning rate has a direct impact on the model's accuracy. An optimal value of $q$ depends on many factors such as: nature of the video (highly-dynamic videos may require lower pruning rates, while mostly static videos may leverage higher pruning rates), desired accuracy/speedup tradeoff, amount of GPU memory available for inference (As we reduce number of input tokens that LLM has to process, peak memory consumption drops leaving more memory for a KV-cache), etc.

Because $q$ can be chosen at inference time, practitioners can dial the compute–accuracy curve on the fly. For example, run at $q=0$ for offline evaluation and $q=0.95$ on resource-constrained edge devices, without further training.

Another thing to consider is the FLOPS ratio of the Vision Encoder and the LLM parts of the VLM. For small models (Qwen-2.5 7B), the overall TTFT speedup effect will be lower than for larger models. The reason is that in the case of small models, the vision encoder takes a considerable part of the compute (see Table \ref{tab:ttft-benchmarking-table}). As LLM size increases, LLM prefill takes an increasingly larger part of it, and larger models benefit more from EVS. It can be seen in Figure \ref{fig:pruning_ablation_uptrain_vs_plugin} (b), where for $q=0.8$, the difference in overall VLM TTFT speedup between Qwen 2.5B 7B and Qwen 2.5 14B is 50\% (191\% and 245\% accordingly).
Regarding the LLM part of the model, we observe a linear dependency between TTFT and the number of input tokens, which remains consistent regardless of model size.

\begin{figure}[ht]
    \centering
    \begin{subfigure}[b]{0.49\textwidth}
        \includegraphics[width=\textwidth]{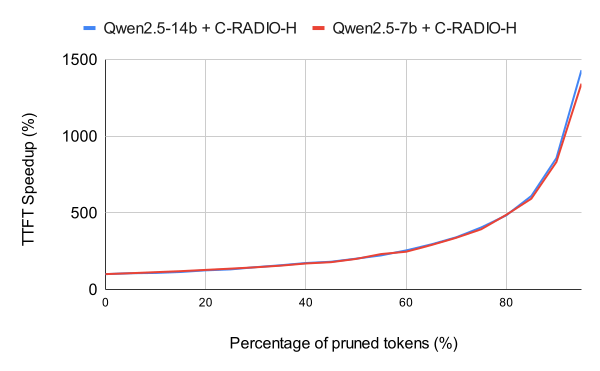}
        \caption{LLM TTFT Speedup}
        \label{fig:llm-speedup-vs-pruning-rate}
    \end{subfigure}
    \hfill
    \begin{subfigure}[b]{0.49\textwidth}
        \includegraphics[width=\textwidth]{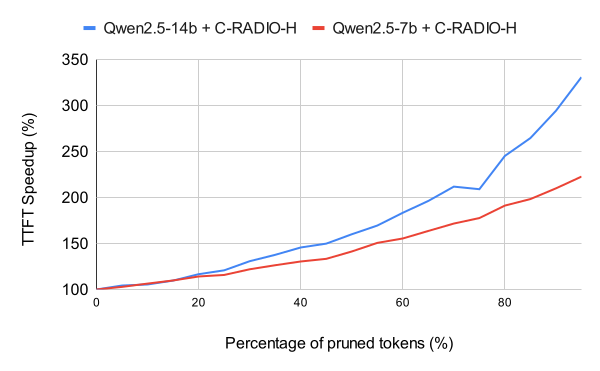}
        \caption{VLM TTFT Speedup}
        \label{fig:vlm-speedup-vs-pruning-rate}
    \end{subfigure}
    \caption{Measured TTFT Speedup for different pruning rates. We measure and report TTFT speedup for the LLM backbone (a), as our method mainly focuses on reducing the number of input tokens into the LLM backbone. We report the overall VLM speedup (b) for completeness.}
    \label{fig:speedup-vs-pruning-rate}
\end{figure}

\begin{figure}[h]
    \centering
    \includegraphics[width=0.49\textwidth]{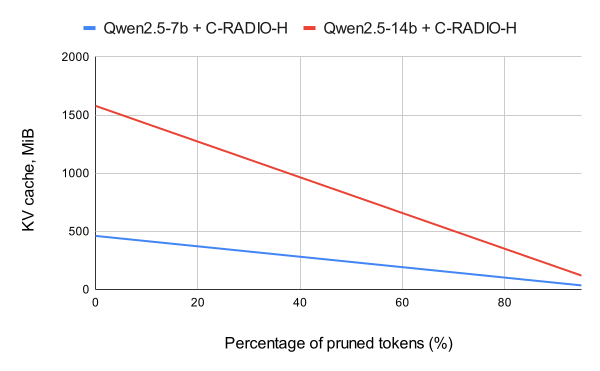}
    \caption{Calculated LLM KV-cache size reduction for different pruning rates.}
    \label{fig:llm-kvcache-vs-pruning-rate}
\end{figure}

\textbf{KV-Cache Memory Reduction.}
In addition to latency and throughput benefits, EVS also provides linear memory savings in the LLM's KV-cache. Since the number of tokens passed to the LLM is reduced in proportion to the pruning rate $q$, the total memory consumed by the attention mechanism—dominated by the KV-cache—also shrinks linearly (Figure \ref{fig:llm-kvcache-vs-pruning-rate}). This effect can be particularly beneficial in multi-stream inference or memory-constrained environments, where minimizing memory overhead is crucial for scalability and batch size. For instance, pruning $q$ of the tokens directly results in $q$ reduction in KV-cache memory usage for the pruned portion, enabling either more concurrent sequences or a larger working set for long-context inference.

Additional benchmark results and KV-cache size derivations can be found in Table \ref{tab:ttft-benchmarking-table} under the Appendix section and Appendix \ref{subsec:kvcache-vs-pruning}.

\section{Experiments}

We conduct comprehensive experiments across multiple video understanding benchmarks to evaluate the effectiveness of EVS. Our setup highlights both the plug-and-play compatibility of EVS and the performance of the uptrained model under varying compression ratios.

\textbf{Datasets.} We evaluate EVS on a diverse set of video understanding benchmarks: VideoMME \cite{fu2025videommefirstevercomprehensiveevaluation} for comprehensive video analysis, MVBench \cite{li2024mvbenchcomprehensivemultimodalvideo} for multi-aspect video understanding, TempCompass \cite{liu2024tempcompassvideollmsreally} for temporal reasoning (We evaluate and report on \textit{MCQ subset} of TempCompass), and nv-Metropolis for activity-centric question answering. These benchmarks encompass a range of video lengths, domains, and reasoning requirements, providing a comprehensive evaluation of EVS performance.

\textbf{nv-Metropolis}, an internal evaluation-only dataset for video question answering. It comprises 331 real-world HD videos, each ranging from 5 to 60 seconds in length, spanning diverse environments such as indoor scenes, outdoor cityscapes, warehouses, healthcare settings, and entertainment venues. Each video is accompanied by multiple human-authored multiple-choice questions (MCQs), totaling 1,900 questions in the dataset. All videos are captured with a single moving camera per scene. The benchmark focuses on accuracy-based evaluation and is designed to assess general-purpose video understanding without requiring audio or external modalities.

\textbf{Baseline Models.} We use Qwen 2.5 7B \cite{qwen2025qwen25technicalreport} and CRADIO-H \cite{Ranzinger_2024_CVPR} vision encoder as our architecture for ablations, configured with eight frames of size $512\times512$ pixels as the standard input configuration. This model serves as a strong baseline for video understanding tasks, allowing us to isolate the impact of our proposed EVS method. For our ablations, we increase the maximum number of frames per video to 8, 32, and 128. 

\textbf{Benchmark Setup.} For the latency benchmarks, we additionally use a larger Qwen 2.5 14B with the CRADIO-H model (Figure \ref{fig:speedup-vs-pruning-rate}). We evaluate inference performance using synthetic data to ensure complete control over input characteristics and reproducibility. All measurements were obtained on an NVIDIA H100 80GB GPU with TensorRT-LLM compilation. Both baseline models were tested with batch size 1, synthetic prompt inputs of 100 tokens, and forced decoding of 128 output tokens. Video input consisted of synthetic tensors representing 32 frames of $512 \times 512$ pixels. The number of image tokens varies based on the EVS pruning percentage.

To measure responsiveness and decoding efficiency, we report two timing metrics: TTFT\textsubscript{llm}, capturing the latency of the LLM component alone, and TTFT\textsubscript{vlm}, which includes the full VLM pipeline. These metrics were obtained using three warm-up runs and three benchmark runs, with the median value reported across all runs. Timing was measured using high-precision counters with proper GPU synchronization. Full benchmark setup and numeric figures are located in Appendix \ref{subsec:runtime-vs-pruning}.

\subsection{Plug-and-Play Evaluation}

\begin{table}[t]
\centering
\begin{tabular}{@{}lccccc@{}}
\toprule
\textbf{Benchmark} 
& \textbf{Baseline}
& \multicolumn{2}{c}{\textbf{Plug-In}} 
& \multicolumn{2}{c}{\textbf{Uptrain}} \\
\cmidrule(lr){3-4} \cmidrule(lr){5-6}
& No pruning & Sequential & Position-preserving & Sequential & Position-preserving \\
\midrule
\textbf{Video MME} $\uparrow$     & & & & & \\
\quad 8 frames                    & 59.90 & 55.30 & \textbf{56.00} & 56.90& \textbf{57.00}\\
\quad 32 frames                   & 65.50 & \textbf{61.40} & 61.30 & \textbf{64.80}& 64.70\\
\quad 128 frames                  & 67.30 & \textbf{65.30} & 64.20 & 68.00& \textbf{68.10}\\
\textbf{nv-Metropolis} $\uparrow$ &       &       &       &       &       \\
\quad 8 frames                    & 74.62 & \textbf{73.85} & 73.76 & 74.26& \textbf{74.17} \\
\quad 32 frames                   & 75.95 & \textbf{75.49} & 74.99 & 75.99& \textbf{76.08} \\
\textbf{TempCompass} $\uparrow$   &       &       &       &       &       \\
\quad 8 frames                    & 69.18 & \textbf{67.91} & 67.59 & 68.35& \textbf{68.73}\\
\quad 32 frames                   & 69.94 & \textbf{69.43} & 68.80 & 69.68& \textbf{70.63}\\
\textbf{MVBench} $\uparrow$       &       &       &       &       &       \\
\quad 8 frames                    & 69.10 & 64.95 & \textbf{66.07} & 66.17& \textbf{67.15}\\
\quad 32 frames                   & 71.80 & 68.97 & \textbf{69.08} & 69.65& \textbf{70.85}\\
\bottomrule
\end{tabular}
\caption{Ablation of position IDs handling, for both plug-in and uptraining variants of use of EVS. Evaluation performed with fixed pruning rate $q=0.75$. We mark scores as \textbf{bold} to indicate a winner in a group: we compare Plug-In EVS independently to Uptrain EVS as these are two distinct models.}

\label{tab:evs-position-id-ablation}
\end{table}

Our primary experiment examines how varying pruning rates impact evaluation metrics when EVS is applied to off-the-shelf models without any additional training. We take pre-trained Qwen models and directly apply EVS at inference time with varying pruning rates $q \in \{0.5..0.95\}$, where higher values correspond to more aggressive pruning (keeping fewer tokens). This experiment demonstrates the immediate applicability of EVS to existing models and quantifies the trade-off between accuracy and efficiency across different compression levels. The results, presented in Table~\ref{tab:evs-pruning-rates-ablation}, demonstrate how model performance degrades or remains stable as the pruning aggressiveness increases, providing insights into the optimal operating points for different use cases.

We also compare plug-in EVS against alternative token reduction strategies, including random pruning, token merging (ToMe), and frame subsampling. As shown in Table~\ref{tab:evs-vs-baselines}, EVS consistently outperforms these baselines across diverse benchmarks. Random pruning yields unstable performance and often discards informative regions, while frame subsampling loses important temporal cues, leading to larger accuracy drops. Token merging performs competitively in some cases but remains less effective overall than EVS, which better preserves spatiotemporal saliency. These results highlight that EVS is not only simple and plug-and-play but also more robust and effective compared to commonly used pruning and compression techniques.

\begin{table}[t]
\centering
\begin{tabular}{@{}lcccc@{}}
\toprule
\textbf{Benchmark} & \textbf{Random} & \textbf{Frame Subsample} & \textbf{Token Merging} & \textbf{EVS} \\
\midrule
\textbf{VideoMME} $\uparrow$ & & & & \\
\quad 8 frames   & 55.80 & 53.40 & \textbf{57.30} & 55.30 \\
\quad 32 frames  & 62.40 & 59.90 & \textbf{62.60} & 61.40 \\
\quad 128 frames & 66.00 & 64.90 & \textbf{66.20} & 65.30 \\
\textbf{nv-Metropolis} $\uparrow$ & & & & \\
\quad 8 frames  & 72.75 & 71.29 & 72.02 & \textbf{73.98} \\
\quad 32 frames  & 75.44 & 74.53 & 74.39 & \textbf{75.55} \\
\textbf{MVBench} $\uparrow$ & & & & \\
\quad 8 frames   & 62.12 & 59.95 & 63.73 & \textbf{64.95} \\
\quad 32 frames  & 65.33 & 68.47 & 68.08 & \textbf{68.97} \\
\textbf{TempCompass MCQ} $\uparrow$ & & & & \\
\quad 8 frames   & \textbf{68.29} & 64.62 & 65.05 & 67.91 \\
\quad 32 frames  & 66.64 & 69.37 & 66.89 & \textbf{69.43} \\
\bottomrule
\end{tabular}
\caption{Comparison of plug-in EVS against baseline token reduction methods (random pruning, frame subsampling, and token merging).
All reduction methods were set to reduce 75\% of tokens. Bold numbers indicate the best-performing method per setting.}
\label{tab:evs-vs-baselines}
\end{table}

\subsection{On Importance of Retaining Position IDs}
\label{sec:exp-position-ids} 

An important design choice in EVS is whether to preserve the original positional information of retained tokens or to treat the pruned sequence as a continuous stream, as described in Section \ref{sec:position-ids}. While preserving original position IDs may seem intuitive for maintaining spatial-temporal relationships, this choice is not apparent because language models are typically trained on dense, consecutive position sequences. When EVS creates sparse position patterns, the model encounters position IDs it was never explicitly trained to handle. This raises the question of whether the model can effectively utilize sparse positional information or if it would perform better with a continuous, re-indexed sequence.

We conduct an ablation study comparing two approaches: (1) \textbf{Position-preserving}, where each kept token retains its original position ID as if it were part of the complete unpruned sequence, and (2) \textbf{Sequential}, where position IDs are reassigned consecutively starting from zero, treating the pruned tokens as a new continuous sequence.

Our results Table \ref{tab:evs-position-id-ablation} reveal an important distinction between plug-in and uptraining scenarios for position ID handling. When applying EVS without uptraining (plug-in), the performance of the sequential and position-preserving approaches is very similar, with sequential reindexing showing slight advantages in most cases. However, with uptraining, the model performs significantly better when using position-preserving encoding. Specifically, with uptraining, position-preserving outperforms sequential in 7 out of 8 evaluated cases, with only Video MME at 8 frames showing a slight underperformance. This suggests that while language models trained on dense, consecutive position sequences initially struggle with sparse position patterns, they can effectively learn to utilize preserved positional information through targeted training. The uptraining process appears to teach the model to leverage spatial-temporal positional cues more effectively, making position-preserving strategies the preferred choice for production deployments.

\subsection{Uptraining with EVS}

To further improve model performance when using EVS, we experiment with an Uptraining phase and evaluate its impact on the resulting model's accuracy. Our uptraining strategy employs a \textbf{Stochastic pruning rate}, where we sample the pruning rate from a distribution, enabling the model to experience varying levels of token compression during training. We ablate it against a \textbf{Constant pruning rate}, where we fix the pruning rate at $q=0.75$ throughout training. 

The stochastic pruning approach exposes the model to a continuum of compression ratios during training, teaching it to be robust across different pruning as well as for no pruning. This is implemented by sampling $q$ from $\text{Beta}(\alpha, \beta)$ where the parameters are chosen such that the mode is approximately 0.75. Still, the model encounters both more aggressive and more conservative pruning rates during training. This variability during training proves crucial for maintaining performance flexibility at inference time.

\begin{table}[t]
\centering
\begin{tabular}{@{}lcccc}
\toprule
\textbf{Benchmark} & \textbf{Baseline} & \textbf{No Uptrain} & \textbf{Fixed Uptrain} & \textbf{Stochastic Uptrain} \\
\midrule
\textbf{Video MME} $\uparrow$ & & & & \\
\quad 8 frames & 59.90& 55.30 & 57.50 & 58.30 \\
\quad 32 frames & 65.50 & 61.40 & 63.60 & 64.70 \\
\quad 128 frames & 67.30 & 65.30 & 66.70 & 68.10\\
\textbf{nv-Metropolis} $\uparrow$ & & & & \\
\quad 8 frames  & 74.62& 73.98 & 74.26& 74.99\\
\quad 32 frames & 75.95& 75.55 & 75.67& 76.36\\
\textbf{TempCompass} $\uparrow$ &  &  &  &  \\
\quad 8 frames  & 69.18& 67.91& 69.11   & 68.73\\
\quad 32 frames & 69.94& 69.43& 70.38& 70.63\\
\textbf{MVBench} $\uparrow$ &  &  &  &  \\
\quad 8 frames  & 69.10& 64.95& 67.75 & 67.15\\
\quad 32 frames & 71.80& 68.97& 70.80& 70.85\\
\bottomrule
\end{tabular}
\caption{Ablation on Uptraining methods. Baseline uses all tokens (no pruning). ``No Uptrain'' stands for Plug-In EVS, ``Fixed Uptrain'' uses constant pruning rate $q=0.75$ and ``Stochastic Uptrain'' samples $q$ from beta distribution during \emph{Uptraining} phase. Pruning is done based on feature-level similarity and preserving position IDs.}
\label{tab:evs-uptrain-ablation}
\end{table}

Our results (Table \ref{tab:evs-uptrain-ablation}) demonstrate that Uptraining with stochastic pruning rates significantly outperforms both the constant pruning rate approach and the plug-and-play baseline. The stochastic approach not only improves final accuracy but also enables the model to perform well under user-defined pruning rates at inference time. This flexibility allows practitioners to adjust the compute-accuracy trade-off on the fly without requiring separate model checkpoints for different operating points. A model trained with stochastic pruning can seamlessly operate at $q=0.9$ (13$\times$ TTFT reduction of LLM) for maximum efficiency or $q=0.5$ for maximum accuracy (2$\times$ TTFT reduction of LLM), making it highly adaptable to different deployment scenarios and computational constraints (See Figure \ref{fig:pruning_ablation_uptrain_vs_plugin} and Table \ref{tab:ttft-benchmarking-table}).

\section{Conclusion}

We have presented Efficient Video Sampling (EVS), a simple yet effective method for reducing token redundancy in video-language models through temporal patch compression. By leveraging the inherent redundancy in consecutive video frames, EVS identifies and aggregates static visual patches while preserving their positional information, enabling significant computational savings without sacrificing model performance.

Our key contributions demonstrate EVS's practical value across multiple dimensions. First, EVS can operate as a training-free, plug-and-play solution that can be immediately applied to any existing vision-language model without architectural modifications or retraining. This immediate applicability makes EVS particularly valuable for practitioners working with pre-trained models in production environments. Second, EVS offers flexible trade-offs between efficiency and accuracy, allowing users to either reduce TTFT and FLOPS while maintaining equivalent accuracy or process longer video sequences and higher sampling rates within the same computational budget. Third, the method introduces negligible computational overhead, requires no additional trained parameters, and can be applied to any VLM architecture.  

From a deployment perspective, EVS's integration requirements are remarkably modest, requiring only two additional gather operations that can be efficiently implemented within existing frameworks, such as TensorRT-LLM. This simplicity, combined with the method's universal compatibility across different VLM architectures and model sizes, positions EVS as a broadly applicable solution for the video understanding community.

EVS addresses a fundamental bottleneck in video-language understanding—the quadratic scaling of computational cost with sequence length—through a principled yet practical approach. By making long-form video processing feasible within existing computational constraints, EVS opens new possibilities for real-world applications in surveillance, instructional video analysis, and continuous robot learning, where processing efficiency directly translates to practical deployment viability.

\subsection{Future Work}

EVS provides a practical and efficient solution for scalable video-language understanding, but several directions remain for future research:

\textbf{Online Video Streams.} In real-time applications such as robotics and physical AI, future work could explore EVS in streaming settings—sampling full keyframes at fixed intervals or based on content dynamics, then pruning intermediate frames using EVS. Combining this with KV-Cache mechanisms to store and reuse tokenized keyframes could significantly accelerate inference and improve the responsiveness of continuous video input. Another potential optimization involves applying non-trivial attention masks to mask out pruned parts of the image within the vision encoder itself. In our current implementation, we pass the entire frame through the vision encoder and only perform masked selection before sending input to the language model.

\textbf{Query-Aware Pruning.} Incorporating task- or language-guided signals into EVS could enable dynamic, query-conditioned token retention, focusing compute only on regions relevant to the user's question or instruction. This aligns EVS with query-driven video understanding, potentially yielding further gains in efficiency without compromising task accuracy.

\textbf{Joint Mask Prediction with Vision Encoder.} Integrating EVS directly into the vision encoder—e.g., predicting pruning masks as an intermediate output—could enable early exiting and minimize redundant computation. This tighter coupling would further reduce TTFT and memory usage, pushing EVS toward real-time performance even on long or high-resolution videos. Moreover, by allowing the encoder to condition mask prediction on the scene's dynamics, EVS can more effectively adapt to varying motion patterns and selectively retain information in regions of interest.

\textbf{Long-Context Modeling.} As EVS reduces token growth to sub-linear levels, it opens the door to processing significantly longer videos than previously feasible. Future work could explore how EVS interacts with emerging long-context LLMs and whether temporal sparsity patterns can be optimized jointly with language attention mechanisms to enhance temporal reasoning across tens of thousands of frames.

By pursuing these directions, EVS could evolve into a fully adaptive, real-time video compression and understanding module, broadening its impact while preserving its core strengths: simplicity, compatibility, and efficiency.

\newpage
\bibliography{refs}

\begin{thebibliography}{10}

\bibitem{alayrac2022flamingo}
Jean-Baptiste Alayrac, Jeff Donahue, Pauline Luc, Antoine Miech, Iain Barr, Yana Hasson, Karel Lenc, Arthur Mensch, Katherine Millican, Malcolm Reynolds, et~al.
\newblock Flamingo: a visual language model for few-shot learning.
\newblock {\em Advances in neural information processing systems}, 35:23716--23736, 2022.

\bibitem{bolya2023tokenmergingvitfaster}
Daniel Bolya, Cheng-Yang Fu, Xiaoliang Dai, Peizhao Zhang, Christoph Feichtenhofer, and Judy Hoffman.
\newblock Token merging: Your vit but faster, 2023.

\bibitem{chen2024internvl}
Zhe Chen, Jiannan Wu, Wenhai Wang, Weijie Su, Guo Chen, Sen Xing, Muyan Zhong, Qinglong Zhang, Xizhou Zhu, Lewei Lu, et~al.
\newblock Internvl: Scaling up vision foundation models and aligning for generic visual-linguistic tasks.
\newblock In {\em Proceedings of the IEEE/CVF conference on computer vision and pattern recognition}, pages 24185--24198, 2024.

\bibitem{cheng2025scalingvideolanguagemodels10k}
Chuanqi Cheng, Jian Guan, Wei Wu, and Rui Yan.
\newblock Scaling video-language models to 10k frames via hierarchical differential distillation, 2025.

\bibitem{choudhury2024dontlooktwicefaster}
Rohan Choudhury, Guanglei Zhu, Sihan Liu, Koichiro Niinuma, Kris~M. Kitani, and László Jeni.
\newblock Don't look twice: Faster video transformers with run-length tokenization, 2024.

\bibitem{fu2025videommefirstevercomprehensiveevaluation}
Chaoyou Fu, Yuhan Dai, Yongdong Luo, Lei Li, Shuhuai Ren, Renrui Zhang, Zihan Wang, Chenyu Zhou, Yunhang Shen, Mengdan Zhang, Peixian Chen, Yanwei Li, Shaohui Lin, Sirui Zhao, Ke~Li, Tong Xu, Xiawu Zheng, Enhong Chen, Caifeng Shan, Ran He, and Xing Sun.
\newblock Video-mme: The first-ever comprehensive evaluation benchmark of multi-modal llms in video analysis, 2025.

\bibitem{hong2025context}
Kelly Hong, Anton Troynikov, and Jeff Huber.
\newblock Context rot: How increasing input tokens impacts llm performance.
\newblock Technical report, Chroma, July 2025.

\bibitem{li2024mvbenchcomprehensivemultimodalvideo}
Kunchang Li, Yali Wang, Yinan He, Yizhuo Li, Yi~Wang, Yi~Liu, Zun Wang, Jilan Xu, Guo Chen, Ping Luo, Limin Wang, and Yu~Qiao.
\newblock Mvbench: A comprehensive multi-modal video understanding benchmark, 2024.

\bibitem{lin2024vila}
Ji~Lin, Hongxu Yin, Wei Ping, Pavlo Molchanov, Mohammad Shoeybi, and Song Han.
\newblock Vila: On pre-training for visual language models.
\newblock In {\em Proceedings of the IEEE/CVF conference on computer vision and pattern recognition}, pages 26689--26699, 2024.

\bibitem{liu2023visual}
Haotian Liu, Chunyuan Li, Qingyang Wu, and Yong~Jae Lee.
\newblock Visual instruction tuning.
\newblock {\em Advances in neural information processing systems}, 36:34892--34916, 2023.

\bibitem{liu2024tempcompassvideollmsreally}
Yuanxin Liu, Shicheng Li, Yi~Liu, Yuxiang Wang, Shuhuai Ren, Lei Li, Sishuo Chen, Xu~Sun, and Lu~Hou.
\newblock Tempcompass: Do video llms really understand videos?, 2024.

\bibitem{liu2025keyframeorientedvisiontokenpruning}
Yudong Liu, Jingwei Sun, Yueqian Lin, Jingyang Zhang, Ming Yin, Qinsi Wang, Jianyi Zhang, Hai Li, and Yiran Chen.
\newblock Keyframe-oriented vision token pruning: Enhancing efficiency of large vision language models on long-form video processing, 2025.

\bibitem{liu2025nvilaefficientfrontiervisual}
Zhijian Liu, Ligeng Zhu, Baifeng Shi, Zhuoyang Zhang, Yuming Lou, Shang Yang, Haocheng Xi, Shiyi Cao, Yuxian Gu, Dacheng Li, Xiuyu Li, Yunhao Fang, Yukang Chen, Cheng-Yu Hsieh, De-An Huang, An-Chieh Cheng, Vishwesh Nath, Jinyi Hu, Sifei Liu, Ranjay Krishna, Daguang Xu, Xiaolong Wang, Pavlo Molchanov, Jan Kautz, Hongxu Yin, Song Han, and Yao Lu.
\newblock Nvila: Efficient frontier visual language models, 2025.

\bibitem{Maaz2023VideoChatGPT}
Muhammad Maaz, Hanoona Rasheed, Salman Khan, and Fahad~Shahbaz Khan.
\newblock Video-chatgpt: Towards detailed video understanding via large vision and language models.
\newblock In {\em Proceedings of the 62nd Annual Meeting of the Association for Computational Linguistics (ACL 2024)}, 2024.

\bibitem{qwen2025qwen25technicalreport}
Qwen, :, An~Yang, Baosong Yang, Beichen Zhang, Binyuan Hui, Bo~Zheng, Bowen Yu, Chengyuan Li, Dayiheng Liu, Fei Huang, Haoran Wei, Huan Lin, Jian Yang, Jianhong Tu, Jianwei Zhang, Jianxin Yang, Jiaxi Yang, Jingren Zhou, Junyang Lin, Kai Dang, Keming Lu, Keqin Bao, Kexin Yang, Le~Yu, Mei Li, Mingfeng Xue, Pei Zhang, Qin Zhu, Rui Men, Runji Lin, Tianhao Li, Tianyi Tang, Tingyu Xia, Xingzhang Ren, Xuancheng Ren, Yang Fan, Yang Su, Yichang Zhang, Yu~Wan, Yuqiong Liu, Zeyu Cui, Zhenru Zhang, and Zihan Qiu.
\newblock Qwen2.5 technical report, 2025.

\bibitem{Ranzinger_2024_CVPR}
Mike Ranzinger, Greg Heinrich, Jan Kautz, and Pavlo Molchanov.
\newblock Am-radio: Agglomerative vision foundation model reduce all domains into one.
\newblock In {\em Proceedings of the IEEE/CVF Conference on Computer Vision and Pattern Recognition (CVPR)}, pages 12490--12500, June 2024.

\bibitem{shen2024longvuspatiotemporaladaptivecompression}
Xiaoqian Shen, Yunyang Xiong, Changsheng Zhao, Lemeng Wu, Jun Chen, Chenchen Zhu, Zechun Liu, Fanyi Xiao, Balakrishnan Varadarajan, Florian Bordes, Zhuang Liu, Hu~Xu, Hyunwoo~J. Kim, Bilge Soran, Raghuraman Krishnamoorthi, Mohamed Elhoseiny, and Vikas Chandra.
\newblock Longvu: Spatiotemporal adaptive compression for long video-language understanding, 2024.

\bibitem{solari1997digital}
Stephen~J Solari.
\newblock {\em Digital video and audio compression}.
\newblock McGraw-Hill Professional, 1997.

\bibitem{wiegand2003overview}
Thomas Wiegand, Gary~J Sullivan, Gisle Bjontegaard, and Ajay Luthra.
\newblock Overview of the h. 264/avc video coding standard.
\newblock {\em IEEE Transactions on circuits and systems for video technology}, 13(7):560--576, 2003.

\bibitem{zhang2023video}
Hang Zhang, Xin Li, and Lidong Bing.
\newblock Video-llama: An instruction-tuned audio-visual language model for video understanding.
\newblock {\em arXiv preprint arXiv:2306.02858}, 2023.

\bibitem{zhang2025llavaminiefficientimagevideo}
Shaolei Zhang, Qingkai Fang, Zhe Yang, and Yang Feng.
\newblock Llava-mini: Efficient image and video large multimodal models with one vision token, 2025.

\bibitem{zhang2025sparsevlmvisualtokensparsification}
Yuan Zhang, Chun-Kai Fan, Junpeng Ma, Wenzhao Zheng, Tao Huang, Kuan Cheng, Denis Gudovskiy, Tomoyuki Okuno, Yohei Nakata, Kurt Keutzer, and Shanghang Zhang.
\newblock Sparsevlm: Visual token sparsification for efficient vision-language model inference, 2025.

\end{thebibliography}

\appendix
\newpage

\section{Additional Results}
\label{app:additional_results}

\subsection{Accuracy loss vs. Pruning rate}

To evaluate how pruning rate affects performance, we perform an ablation over a range of pruning thresholds $q \in \{0.5..0.95\}$. We also compare the effect of uptraining a model compared to the plug-and-play EVS approach.

\begin{table}[htbp]
\centering
\begin{tabular}{@{}lcccccccc@{}}
\toprule
\textbf{Pruning rate}
& \multicolumn{2}{c}{\textbf{Video MME} $\uparrow$} 
& \multicolumn{2}{c}{\textbf{nv-Metropolis} $\uparrow$} 
& \multicolumn{2}{c}{\textbf{TempCompass} $\uparrow$} 
& \multicolumn{2}{c}{\textbf{MVBench} $\uparrow$} \\
\cmidrule(lr){2-3} \cmidrule(lr){4-5} \cmidrule(lr){6-7} \cmidrule(lr){8-9}
& \textit{Plug-in} & \textit{Uptrain} 
& \textit{Plug-in} & \textit{Uptrain} 
& \textit{Plug-in} & \textit{Uptrain} 
& \textit{Plug-in} & \textit{Uptrain} \\
\midrule
No pruning   
& \multicolumn{2}{c}{65.50}                
& \multicolumn{2}{c}{75.95}              
& \multicolumn{2}{c}{69.94}                 
& \multicolumn{2}{c}{71.80} \\

$q=0.50$     
& -1.55\% & -1.08\%  
& -0.30\% & +0.24\% 
& -0.46\% & +0.54\% 
& -2.24\% & -1.84\% \\

$q=0.55$     
& -2.50\% & -1.24\%  
& -0.24\% & -0.12\% 
& -0.37\%  & +0.44\% 
& -2.53\% & -1.73\% \\

$q=0.60$     
& -3.15\% & -1.39\%  
& -0.42\% & +0.06\% 
& -0.46\% & +0.63\% 
& -2.75\% & -1.63\% \\

$q=0.65$     
& -3.64\% & -1.39\%  
& -0.79\% & +0.06\% 
& -0.46\% & +0.63\% 
& -3.19\% & -1.63\% \\

$q=0.70$     
& -5.31\% & -1.71\%  
& -0.91\% & -0.12\% 
& -1.38\% & +0.63\% 
& -3.68\% & -1.60\% \\

$q=0.75$     
& -6.85\% & -2.83\%  
& -1.28\% & +0.18\% 
& -1.66\% & +0.81\% 
& -3.94\%  & -2.06\% \\

$q=0.80$     
& -7.55\% & -2.99\%  
& -1.84\% & -0.06\% 
& -2.51\% & +0.71\% 
& -4.41\% & -1.77\% \\

$q=0.85$     
& -9.72\% & -5.31\%  
& -1.71\% & -0.97\% 
& -2.79\% & -0.37\% 
& -5.05\% & -2.50\% \\

$q=0.90$     
& -13.72\% & -6.50\%  
& -2.34\% & -2.21\% 
& -4.54\% & -0.10\% 
& -7.65\% & -3.53\% \\

$q=0.95$     
& -20.40\% & -11.02\% 
& -3.87\% & -3.10\% 
& -7.92\% & -1.20\% 
& -15.90\% & -9.65\% \\
\bottomrule
\end{tabular}
\caption{
Impact of pruning rate on accuracy across benchmarks. Pruning is applied post-encoder on feature-level embeddings, preserving positional IDs. We compare two settings: \textit{Plug-in EVS} and \textit{Uptrained EVS}. Values are shown as percentage change are relative to the no-pruning baseline.
}
\label{tab:evs-pruning-rates-ablation}
\end{table}

\subsection{Runtime numbers vs. Pruning rate}
\label{subsec:runtime-vs-pruning}

All benchmarks were conducted using synthetic data on a machine equipped with an NVIDIA H100 80GB HBM3 GPU (driver version 535.216.03, CUDA 12.9) and dual-socket Intel(R) Xeon(R) Platinum 8462Y+ CPUs with 128 threads total. The system ran Ubuntu 24.04.2 LTS with Python 3.12.3, PyTorch 2.7.0, and cuDNN 9.9. The total available system memory was approximately 2 TB.

The models were compiled using TensorRT-LLM v0.20.0rc2.
Benchmarks were run with a batch size of 1, using input prompts of 100 synthetic tokens and synthetic video inputs with 32 frames per video, 1 tile per frame, and tile size of $512 \times 512$ pixels. Vision frames were generated as \texttt{torch.float16} tensors of shape $[1, 32, 3, 512, 512]$. Token generation was fixed at 128 output tokens, with one decoding stream.

For each EVS configuration, three warmup runs were followed by three timed runs. The median values from the timed runs were reported.

Two time-to-first-token (TTFT) metrics were captured:
\begin{itemize}
    \item \textbf{TTFT\textsubscript{llm}}: Time from the LLM prefill start to the first token generated, excluding the vision encoder.
    \item \textbf{TTFT\textsubscript{vlm}}: End-to-end TTFT, including the vision encoder and LLM pipeline.
\end{itemize}

All timing measurements used \texttt{time.perf\_counter()} with appropriate GPU synchronization to ensure accurate recording. 

\begin{table}[htbp]
\centering
\begin{tabular}{@{}lcccccc@{}}
\toprule
\textbf{Pruning Rate} 
& \multicolumn{3}{c}{\textbf{Qwen 2.5 7B + CRADIO-H}} 
& \multicolumn{3}{c}{\textbf{Qwen 2.5 14B + CRADIO-H}} \\
\cmidrule(lr){2-4} \cmidrule(lr){5-7}
& TTFT\textsubscript{vlm} & TTFT\textsubscript{llm} & Latency 
& TTFT\textsubscript{vlm} & TTFT\textsubscript{llm} & Latency \\
& (sec) & (sec) & (sec) & (sec) & (sec) & (sec) \\

\midrule
No pruning   & 0.3138 & 0.1892 & 100.07 & 0.50528 & 0.37860 & 101.37 \\
$q=0.05$     & 0.3053 & 0.1786 & 100.85 & 0.48462 & 0.35841 & 102.52 \\
$q=0.10$     & 0.2950 & 0.1679 & 101.79 & 0.47940 & 0.35163 & 103.48 \\
$q=0.15$     & 0.2859 & 0.1586 & 102.47 & 0.46073 & 0.33251 & 103.97 \\
$q=0.20$     & 0.2750 & 0.1484 & 103.70 & 0.43339 & 0.30392 & 105.40 \\
$q=0.25$     & 0.2711 & 0.1393 & 103.92 & 0.41836 & 0.28932 & 106.29 \\
$q=0.30$     & 0.2574 & 0.1309 & 105.37 & 0.38660 & 0.26041 & 107.08 \\
$q=0.30$     & 0.2484 & 0.1218 & 106.34 & 0.36713 & 0.23972 & 108.32 \\
$q=0.40$     & 0.2407 & 0.1117 & 106.91 & 0.34700 & 0.21959 & 109.12 \\
$q=0.45$     & 0.2355 & 0.1064 & 107.61 & 0.33732 & 0.20949 & 110.26 \\
$q=0.50$     & 0.2221 & 0.0950 & 108.97 & 0.31585 & 0.18854 & 111.57 \\
$q=0.55$     & 0.2084 & 0.0820 & 107.50 & 0.29821 & 0.17059 & 113.02 \\
$q=0.60$     & 0.2019 & 0.0769 & 112.23 & 0.27560 & 0.14865 & 114.52 \\
$q=0.65$     & 0.1918 & 0.0654 & 113.29 & 0.25752 & 0.12883 & 115.81 \\
$q=0.70$     & 0.1828 & 0.0561 & 114.43 & 0.23862 & 0.11130 & 117.19 \\
$q=0.75$     & 0.1767 & 0.0482 & 115.17 & 0.24182 & 0.09358 & 115.22 \\
$q=0.80$     & 0.1642 & 0.0389 & 116.62 & 0.20612 & 0.07830 & 119.31 \\
$q=0.85$     & 0.1583 & 0.0320 & 117.39 & 0.19100 & 0.06210 & 120.54 \\
$q=0.90$     & 0.1495 & 0.0228 & 120.34 & 0.17178 & 0.04421 & 123.54 \\
$q=0.95$     & 0.1409 & 0.0141 & 121.61 & 0.15277 & 0.02648 & 124.92 \\
\bottomrule
\end{tabular}
\caption{Benchmark results of various models measuring TTFT speedup of the LLM and entire VLM as a function of pruning rate factor $q$. Models we compiled with TRT-LLM engine. Benchmarking was done with a batch size of 1, 100 input prompt tokens, 32 frames per video, 1 tile per frame, 512px tile size}
\label{tab:ttft-benchmarking-table}
\end{table}

\subsection{KV-cache Memory Calculation}
\label{subsec:kvcache-vs-pruning}
The total memory required for attention-related activations and weights during inference can be broken down into three main components: KV-cache storage and the model’s attention weights. The total number of tokens for which the KV-cache must be allocated is computed as the product of the effective sequence length $S$ and the sum of batch size $B$ and prefill queue size $Q$. The KV-cache dimension per token, denoted $D_{kv}$, depends on the model architecture (e.g., number of heads and embedding size). If query vectors must also be kept during prefill, an additional buffer of size $S \cdot d_{\text{model}}$ is included. The overall memory footprint $M$ in MiB is then given by:

\begin{equation}
M = \frac{1}{2^{20}} \left( (S \cdot (B + Q) \cdot D_{kv}) \cdot s_{\text{kv}} + \delta \cdot (S \cdot d_{\text{model}}) \cdot s_{\text{w}} + P \cdot s_{\text{w}} \right)
\end{equation}

where $s_{\text{kv}}$ and $s_{\text{w}}$ are the byte sizes of the KV-cache and weights data types, respectively, $P$ is the number of attention parameters, and $\delta \in \{0, 1\}$ indicates whether query prefill is allocated. The KV-cache memory alone is given by:

\begin{equation}
M_{\text{kv}} = \frac{1}{2^{20}} \cdot (S \cdot (B + Q) \cdot D_{kv}) \cdot s_{\text{kv}}
\end{equation}

All other factors are independent of sequence length, so the KV-cache memory usage increases linearly with sequence length $S$.

\end{document}